\pdfoutput=1
\documentclass[11pt]{article}
\usepackage{acl}
\usepackage{times}
\usepackage{latexsym}
\usepackage[T1]{fontenc}
\usepackage[utf8]{inputenc}
\usepackage{microtype}
\usepackage{graphicx}

\title{ LLMs and the Human Condition }
 \author{Peter Wallis \\
  pwallis@acm.org}

\begin{document}
\maketitle
\begin{abstract}
This paper proposes a model of what Large Language Models are doing when they impress us with its language skills.  Rather than doing classic AI with ``distributed representations,'' the proposal is that LLMs work because they have been trained on descriptions of {\bf practices}, and ``glorified auto-complete'' is sufficient to capture the mechanism behind a significant proportion of our rather shallow intelligence.  Although not flattering to our modern ego, the approach provides an explanation of apparent ``mind reading'' in human communication -- a phenomena that has been  challenging for conversational user interfaces.

\end{abstract}

For the last few years work in AI has become synonymous with machine learning and in particular neural net architectures for Large Language Models or LLMs.  These systems do impressive things, but \textit{how} they do them is poorly understood.  The idea that we do things by thinking about it is very appealing and, with the growth of interest in AI, many coming from outside the field have not studied the history. Inevitably old mistakes get repeated.
The paper starts with a recap of good old fashioned AI (GOFAI). It then describes what Brooks and many philosophers have said about the nature of situated action. It then introduces a mechanism that was touched on in the 1970s in AI research but which is an idea that is hundreds of years old in what is now sociology.  Finally the paper discusses how the mechanism can be applied to a major issue in Natural Language Understanding (NLU) namely the phenomenon of apparent ``mind reading''.

\section{The problem: people read minds}

Working on NLU, one sooner or later comes across a phenomenon in which humans recognise the intent of their conversational partner (for example \cite{Wallis08}).  The solution is often assumed to be a better (and more detailed) representation of the context and from that perspective this paper proposes representing context, not as things and their relationships, but as \textbf{an ecology of practices}.

A good example of the problem is this transcript from Mann~\cite{Mann88} --- a transcript of a real conversation in a naturally occurring setting:

\begin{center}
\small
\begin{tabular}[h]{rrp{5cm}}
  1&Child:& I'm hungry \\
  2&Mother:& Did you do a good job of your geography homework? \\
  3&Child:& Yeah. \\
  4&Child:& What's to eat? \\
  5&Mother:& Let me read it. \\
  6&Mother:& What is the capital of Brazil? \\
  7&Child:& Rio de Janeiro. \\
  8&Mother:& Think about it. \\
  9&Child:& It's Brasilia. \\
  10&Child:& Can I eat now? \\
  11&Mother:& I'll let you have something later. \\
  12&Mother:& What is the capital of Venezuela? \\
  13&Child:& Caracas. \\
  14&Mother:& Fine. \\
  15&Child:& So what can I eat? \\
  16&Mother:& You want some cereal? \\
  17&Child:& Sure. \\
\end{tabular}
\end{center}

\noindent
All English speakers \textit{by definition} will follow what is happening in this conversation and feel there is nothing extraordinary about it.  The problem for a reference model of language understanding is that the mother says nothing related to the child's utterances until line 11. There is no overlap of semantic content - none at all.  Of course, as English speakers, we can all say what both the mother and the child \textit{want}.

This has historically lead researchers interested in conversational agents off into the realm of recognition of intent and modelling theory-of-mind (ToM)
\cite{trains,lev2006}. Classically this would be modelled in terms of the beliefs, desire and intentions of an agent, the theory being that a rational agent will do what it believes is in its interests~\cite{cohe90}, all in terms of good old fashioned symbolic representations.  But is that how minds actually work?

\section{Artificial Intelligence - a history}

Computers are very good - indeed ideal - at manipulating symbols.  In general, words and symbols are seen as having meaning by \textit{referring} to things in the world, and the early AI researchers saw language understanding as a problem of mapping messy natural languages into something more tidy~\cite{mel81} -- more pure -- as God would have made it~\cite{Eco95}.  For centuries there has been the idea that meaning bottoms out at some kind of semantic primitives \cite{ishLei} and in this tradition Roger Schank proposed that there was a workable set of primitive verbs or `acts' on which the meaning of a sentence could be based~\cite{sch72}.  Examples are PTRANS for ``physical transfer'' and MTRANS for mental or knowledge transfer, and that machines could translate natural descriptions into (representations of) meaning and figure out consequences from there.

The primitives approach to AI didn't really work but there was also the idea that a formal system of symbol manipulations might map isomorphically onto events in the world.  In ``Godel, Escher, Bach'', Hofstader~\cite{Hofstader79} gives a beautiful example with his pq- system which is a set of syntactic transforms on strings of 'p', 'q', and '-'. The system produces an infinite set of strings including ``\verb!-p--q---!'' and 
``\verb!--p--q----!'' but not ``\verb!-p---q--!''. The number of '-'s in  produced strings just happens to map onto the number of apples in a box as you add more, or take them out.  The symbol 'p' seems to mean ``plus'' and 'q', ``equals'' because they appear in the system of productions in an isomorphic relationship with our system of adding and subtracting.  They have no meaning in themselves without the system in which they participate. Rather than meaning being built up by combining primitive features with innate meaning, the ``primitives'' get meaning from the structure in which they occur.

Computers are good at this kind of syntactic symbol manipulation, and between that and the observation that it is hard to think of a thousand things you know a hundred things about, this kind of reasoning formed the basis of thinking about computers and meaning.  Today a spreadsheet is a great way to automate reasoning about some things but it turns out there is a wide range of tasks that fail when handled this way.  Counting apples in a box might seem easy, but a box might have 15 apples in it according to Waitrose, 25 according to Aldi, and a horse would eat the lot no matter how many are rotten.  Modelling the world with referential symbols is hard because the referent can change. If the idea is to map a room before you vacuum it, cats and people tend move and re-set the model.  In the early days of robot football the machines would wait for the ball to stop before deciding what to do. Air traffic control has a similar problem.  At the time of writing aircraft have been grounded while air traffic controllers sort out why an aircraft, represented by a flight number, occurred in two locations. There is something wrong with the system's representation of the state of the world.  There were projects set up to formally encode all the data in the world, the CYC project~\cite{cyc} being the most explicit.  The feeling in the late 1980s was that no matter how powerful computers might become in the future, simply getting the data for symbolic representations was going to be problematic, but perhaps there was a better way.

Rather than using sense-data to form a representation of the world, then reasoning about it before acting, functional systems could be built by connecting sensing to acting. Roomba vacuum cleaners do a random walk of a room to clean it and avoid chairs (and cats) by touching them and turning away.  Since 1990 there has been a glacial shift in the computer science collective understanding of the problem and today there is wide spread acknowledgement~\cite{wooldridge23} that the nature of intelligence is inseparably linked to our embodiment in the world.

How this pans out for machines is still not clear and Brooks' robots are often dismissed as being just ``insect level intelligence'' with robot developers continuing to add symbolic representations in order to support higher level intelligence (for example see \cite{Ark98}).  Philosophers on the other hand continue to push the limits of what higher level behaviours can be explained in terms of reactive systems~\cite{HeJa96, varieties17, gallagher2020, Bar-On2021}.
The notion of a ``reactive system'' describes a mechanism used for AI problems, but it is not an interesting ``algorithms and data-structures'' solution.  The ELIZA mechanism takes a set of if-then rules to map input to output, and applied to the DOCTOR script was found to be quite engaging.  That same mechanism is used today for a myriad of chatbots but at the time and for many years after, AI researchers would consider it just a trick.

From a philosopher's perspective, the system is more interesting when considered as just a part of the agent-environment system.  ELIZA worked in a particular environment in exactly the same way as Roombas work in a particular environment and, the claim is, in exactly the same way we humans make most of our decisions about what to do next.

\section{Natural Intelligence - a theory}

 Deciding if I should buy that bottle of whiskey might involve reasoning with symbols representing money, but getting something to eat when I know where the refrigerator is - that is perhaps literally a no brainer.  The radical enactivists~\cite{varieties17} point out how agents and their environment are often set up so that the environment triggers; the environment ``brings on'', or the agent ``directly perceives'' something that causes the agent to behave in an appropriate way with no need for intermediate representation or indeed thinking.  Like the cat, when hungry I automatically move to the refrigerator where the handle ``calls out'' to me to open the door.  Having opened the door, a packet of gnocchi catches my eye (because I am hungry) and picking up the packet, the pan-handle and tap guide my next actions. Like the insect crossing a pebble beach\footnote{I recall this example from undergraduate lectures but can't find a citation.} the behaviour looks complex. Indeed we might be tempted for whatever reason to ascribe beliefs and desires to the insect, but what unfolds is a ``trajectory'' \textit{produced} by the agent but \textit{directed} by the environment. Looking back on my actions when cooking gnocchi I will of course explain them in terms of beliefs and goals, but in the act of doing them, the mechanism does seem to be reactive.  Unfortunately is is  completely uninteresting to real computer scientists.

We humans are not entirely reactive agents because we can, on occasion, think about things like whether I can afford that whiskey using an expression such as ``£38 - £44 = ...'' but we also have another trick.  We humans modify the environment to fit with us.  In the cooking gnocchi for lunch scenario, none of the things I interact with are naturally occurring -- someone made them. What is more, these things were put where I would find them.  I might use insect level intelligence to eat lunch, but surely others must have had thoughts (with symbols) in order to create our benign environment.  It seems this is not the case.

Recently we have been talking with sociologists who look at human decision making at the macro level.  The strong reductionist position is alive and well in sociology and many take the line that collective action is simply made up from the behaviour of individuals making their own decisions - basically that sociology is just psychology on a massive scale in the same way as drunk physicists will claim chemistry is just physics. Social simulation based on the behaviour of individuals is of course entirely possible with today's computers.  There are however problems with this in that collective action does not fit with rational self interest of individuals. Why did Britain leave the EU?  And why does Germany want to go to war with a nation that 2 years ago was a key trading partner? At the other end of the ``structure and agency'' debate~\cite{w_sna} is the idea that society consists essentially of structures such as institutions, traditions, and norms, and that \textbf{people are merely acting out their role} in those structures. Post The Enlightenment we collectively like to think we can reason about things and make rational decisions that are better than the status quo. And it is of course not very flattering to be told you are merely a cog in a machine, but a key finding of AI research over the last 70 years is of course that rationality is not all it is cracked up to be.

So what is the alternative to rational actors reasoning over representations of the world?  The claim is that the environment -- the physical and social -- is evolved rather than there by design.  Rather than smart people having a meeting and inventing the stock-market, the practices making up the stock-market, like the human eye, evolved from what came before. The stock-market might have been set up by a committee, but concepts such as trade and money already existed, as did the process of collective decision-making by committee. Whatever the mechanism of change, the current institutions have been selected for by evolutionary pressures.  Society is an ``ecology of practices'' where the practices are the thing that survive and people are merely the actuators for roles. As our techie once said, ``Cogito, ergo, I'm a cog.''~\cite{leon}.

It is an \textit{ecology} of practices because the practices fit together in some way to create a viable environment -- the environment in which we live and thrive.  These institutions do not, indeed cannot, stand alone.  My going to the fridge when hungry works only because I was raised in a society where there are supermarkets and I do a weekly shop.  I might, but usually do not, plan to eat gnocchi at one pm next Wednesday, but rather see gnocchi on the shelf and go ``I'll have that''.  Having bought gnocchi and put it in the fridge, when I open the refrigerator door a few days later, there it is and cooking and eating practices take over.

And like an ecology in the natural environment, there is a certain amount of autopoiesis resulting in an ecology of practices being robust.  In the animal kingdom, if one species is suddenly removed a gap appears in the resource/consumer space. The gap is either quickly filled by the remnants of that species, something else comes along to fill the space -- rabbits replace bilbies -- or the  resources are appropriated by neighbours in that space. Reduce the number of cats in Sydney, and brown snakes move in to eat the rats and mice~\cite{SydneySnakes}.  If the local supermarket closes, I do not starve, but engage another no doubt older hunter/gatherer practice. I wander about attentively until I find indicators of something to eat - an Aldi sign perhaps - which triggers the more modern practice of shopping.

The society's practices may or may not be visible to us mere mortals, and of course society's practices may not accord with the survival strategies of individual humans -- war being a classic example.  Social insects like ants and bees do not lay down their lives out of love for their fellow insect, but because they are wired that way. Indeed an individual ant does not (let's assume) have a strategy, or indeed even a practice. Instead the ant society has practices, including attacking intruders. The mechanism for implementing it is to produce ants that perform the role of soldier ants.  As humans we may be tempted to claim that the ant nest has a strategy for protecting itself but that would suggest that the nest in some way has goals and representations of an enemy, war, and casualties. The notion of a practice is that it happens, with or without representations of the world or beliefs, with or without goals or desires, and with or without any sense of formulating an intention.
\begin{figure*}[t]
\includegraphics[height=50mm]{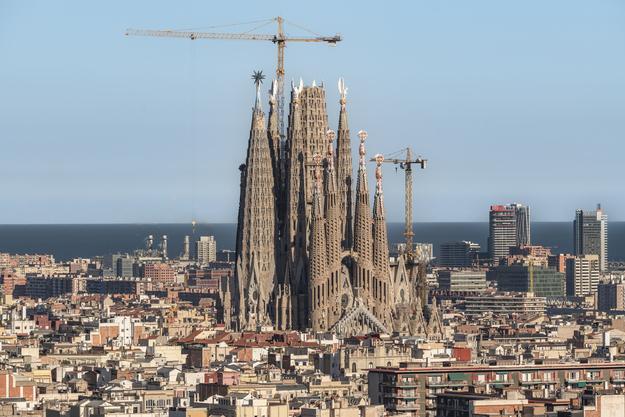}
\includegraphics[height=50mm]{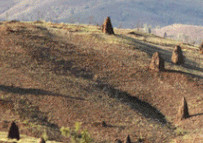}
\caption{Sagrada Familia, Barcelona, vs Termite mounds in Western Australia}
\end{figure*}
Some have drawn parallels between termite mounds and cathedrals and suggested that a termite colony must have some form of ``distributed representation'' that enables (rational) decision making.  Instead, perhaps we should look at Sagrada Familia and think of it as an example of ``insect intelligence'' in humans.

A more positive observation is that we do have the opportunity to do rational choice. And we do it with symbolic representations of things in the world.  Gaudi undoubtedly thought about Sagrada Familia, but he did it in terms of the established practices in which individual stone masons and accountants, priests, engineers and parishioners, all fill roles.  But following the strong version of the linguistic relativism argument, l'sign is arbitraire~\cite{culSau} and what we reason about is culturally constructed.  Gaudi probably thought about bricks at some point, but the concept of a brick~\cite{WgsnBB} was contingent on the practices of stone-masons and brick-layers. It may \textit{seem} obvious that the world is full of things that languages merely name, and that is why languages are largely translatable, but it is proposed that languages ``carve reality''~\cite{wfdt} in largely the same way because human interests are largely shared. The signifieds of human languages are (largely) the same because we interact with the world in a human-like way.  However we humans have a unique skill we have learnt to ``thingifying'' the stuff around us, giving it a label, and then reasoning symbolically with the label. The choice of stuff to label is up to us and based on our ideas about how the world works \textit{as perceived through our successful practices}.

\section{What LLMs are actually doing}

Humans go through life with a set of practices that they know how to do. Rather than sensing the world, forming a representation of the salient parts, and then doing (symbolic) reasoning to produce action that \textit{happens} to string together to form practices,
the basic unit of intelligence as we know it is the practice. Practices as part of an ecology of practices produced by a culture and which facilitates the propagation of the culture.

The proposal is that the things we write down are to a very large extent descriptions of practices. LLM training data is a description of human practice. Sure there are references in there to things, but LLMs don't capture that.  A Generative Pre-trained Transformer -- or any mechanism for generative prediction -- is not ``understanding'' that a cat has four legs, sharp teeth and fur, what it is learning is that we pat cats, we attend when they catch mice, and that they jump on window sills and knock things off. The things we notice about cats are all integrated into our own practices, and what we talk about is those practices.  In order to play chess~\cite{chess,more-chess} LLMs do not represent pieces and strategies; what they do is use ``glorified auto complete''~\cite{GAC} on descriptions of chess games -- on descriptions of the \textit{practice} of playing chess.  It certainly looks like it's reasoning, but that is just anthropomorphism about how we think we solve the problem.

\section{A 'pragmatics first' model of NLU}

We animals behave in accordance with both the reactive framework in which the environment ``causes'' action, and a framework of practices which evolved through evolutionary pressures on our society.  So how does this relate to the mechanism of language as interaction?  Language can of course be used to describe practices in terms of things as described above, but using language is also part of practices.

Returning to the problem of mind reading, the radical enactivists have claimed that in many cases what looks like reasoning about other minds and ToM is actually the ``direct perception'' of another's intent.  (see discussion of Hutto in \cite{gallagher2020}) 
The hypothesis is that an LLM, rather than abstracting from ``cereal'' to ``food'' and then reasoning about ``hunger,'' is generalising from N descriptions of a practice to a rewriting of the practice selected and modified by the current prompt.  Prompted by ``I'm hungry'' an LLM will ``select'' the practice of mother feeding a child - because this is a standard opening for a child to a mother - and fill it in. If the prompt includes the statement that the child likes cereal, then ``want some cereal'' would be a very plausible magic response.

For those who know their history of AI, practices may look much like Schank and Abelson's scripts~\cite{schank89}.  The problem was that scripts as envisaged needed to be hand crafted and relied on symbolic representations of things like restaurants, bills, eating and payment.  The proposal is that LLMs get this data from internet scrapers. Rather than forming representations, they are using ``auto complete'' on the raw data.  The child in our example dialog above can want to eat something and, from past experience know that making the sounds for "I'm hungry" often results in mother feeding him.  We might think that the child means the words he says, and he might, but the act of saying it causes the world to change such that there is no longer a sensation of hunger.

Of course in the example this does not happen.  The child's environment (i.e. Mother) does something different to what is expected at step two. In the child's repertoire of practices however, mother's utterance is a fit with the practice of checking homework. The child then has a decision to make: continue with the practice of getting something to eat, or to switch his Practice in Progress, or PiP, to checking homework.  It is here that the child gets to do ``purposive human action'' -- to reason about what to do next. The thing reasoned about is \textit{not} his choice of words however, but about his choice of practice to perform. This level of decision is not linguistic - it is not about syntax and semantics, but about all sorts of non-linguistic but very pragmatic factors such as the power relation between mother and child.  In making the decision, the child can predict what will happen next by remembering how each practice played out in the past.  Presented with line 11, (``I'll let you have something later.'') the child knows what is coming next if he doesn't stop niggling.  The child does not need to reason about minds -- about why mother is talking of homework -- the child merely has to identify the practice in play and act out the appropriate role. Recognising the PiP is no more complex than recognising the words.  Negotiating the PiP is where the interesting and I propose ``conscious'' decisions are being made.

It is interesting to note that the child's role in the check-homework practice is passive - the child answers questions and waits for the next. When Mother stops asking questions, he (tentatively) reintroduces his own PiP. The suggestion is that, like the earlier example of hunting and gathering, this is a basic, fundamental, or possibly hardwired practice that a human can fall back on that enables learning new practices.  The next time mother says ``have you done a good job of your geography homework?'' he knows what to expect. He will be able to ``read mum's mind''.

\section{Human scale intelligence}

How does a head learn practices? The proposal is that a head does good old fashioned unsupervised learning on the raw data of what actions lead to which opportunities. The world, in combination with the human body's ability to act on it, provides interaction that is consistent enough for regularities to be
detected.  As a consequence, the mind has a catalog of "sequences of interaction" - of practices - which are coherent enough to provide something with a beginning, a middle, and an end. When the agent/world interaction triggers a practice beginning, the agent can predict that, should the middle follow, the end will occur. If the agent can provide that middle, and somehow wants the outcome, then the agent can do what looks like rational action without using symbolic representations of anything.

What the individual does need to do is to learn how to perform roles but that can done without representation.  The individual no doubt can learn by directly experiencing sequences of action and reaction, but many animals, including humans, also seem to experience  the experiences of others (``second person inter-subjectivity'' -- see \cite{gallagher2020}).  We also \textit{imitate} the practices of others (see for example Gergely and Csibra~\cite{GerCsi}).  There are certainly interesting species specific behaviours that enable the training of individuals to perform roles but the key is that these often operate at a very shallow level.

Key to the mechanism as communication is that practices are largely shared by the community of speakers, and that the practices are largely visible to others.  If someone walks up to an ATM machine with a card in their hand, it is fairly likely that they will not start ordering a coffee. This is obvious to us humans -- we `directly perceive,' not just the actions, but also the trajectory of which the actions are the start. And it is not just humans.  A sheep dog does not need to model the mind of a sheep -- the sheep's observable actions are simply a recognisable part of a practice that sheep perform (Figure 2).
\begin{figure*}[ht]
\includegraphics[height=48mm]{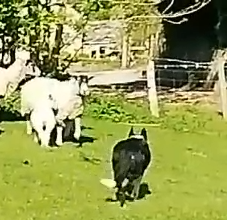}
\includegraphics[height=48mm]{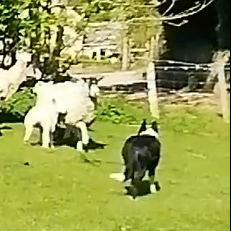}
\includegraphics[height=48mm]{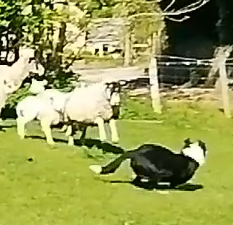}

\caption{Zak doesn't read the Ewe's mind, but instead sees a nasty head but unfolding.}
\end{figure*}

\section{Summary}
In large part the achievements of us humans are a product, not of individuals thinking about stuff, but of societies evolving to do what they do.  A society is a set of practices that are mutually dependent and supportive; human agents are merely the mechanism by which practices are executed. Mostly. Societies compete, and those with the better combination of practices survive. Of course modern humans can think about and change practices, and our super power is to do that with symbols, but the symbols -- critically the signified referenced by the sign -- needs to be identified first. That process is grounded in the practices a society has evolved. Some practices have obvious advantage when looking at the survival of a society. Farming, child rearing and building shelters are fine, but others -- going to church or a football match, elections, and street parties -- are less clear.

At all levels an individual acts a role in different practices at different times, and does not need to be aware or think about why the practice is good for anyone or anything.  Whatever that mechanism, representing practices enables us to predict what happens next. This is how language works. We see a child starting the practice of getting something to eat and we all expect -- including the child -- that the result will be some food. When that doesn't happen, we \textit{could} reason about the contents of the mind of the mother but we might, like the dog above, simply recognise the consequences of a power relationship.

\section{Conclusion}

We don't need to read minds, we just need to predict what comes next and this is what a ``generative pre-trained'' whatever does well when it is trained on what are primarily stories of practices being played out.  For us engineers wanting to build conversational user interfaces, it seems glorified auto complete, implemented with transformers or not, is a useful tool given the right data - the right data being descriptions of practices.
 The way ahead is to find tools for harvesting practices in a way machines can use them. 

\bibliography{../mybib.bib}

\end{document}